\documentclass[twocolumn]{article}
\usepackage{natbib} 
\usepackage{titlesec} 
\usepackage{mathtools}
\usepackage{amsmath}
\usepackage{hyperref}
\usepackage[margin=1.52cm]{geometry} 
\usepackage{enumitem} 
\usepackage{booktabs} 
\usepackage{comment}
\usepackage{tabularx} 
\usepackage{booktabs} 
\usepackage{amsfonts} 
\usepackage{caption}
\usepackage{subcaption} 
\usepackage[most]{tcolorbox}
\usepackage{amsmath}
\usepackage{graphicx}

\titleformat*{\section}{\large\bfseries}
\titleformat*{\subsection}{\normalsize\bfseries}
\titleformat*{\subsubsection}{\normalsize\bfseries}

\title{Chinchilla Scaling: A replication attempt}

\author{Tamay Besiroglu \and Ege Erdil  \and Matthew Barnett  \and Josh You}
\date{%
    \normalsize
    Epoch AI\\%
   \today
}

\begin{document}

\maketitle
\begin{abstract}
Hoffmann et al. (2022) propose three methods for estimating a compute-optimal scaling law. We attempt to replicate their third estimation procedure, which involves fitting a parametric loss function to a reconstruction of data from their plots. However, we find that their reported estimates are inconsistent with their first two estimation methods, fail at fitting the extracted data, and report implausibly narrow confidence intervals—intervals this narrow would require over 600,000 experiments, while they likely only ran fewer than 500. Two factors explain these findings: firstly, the optimizer used by Hoffmann et al. stopped before convergence due to a poor choice of loss scale, and secondly, the parameter estimates reported in the body of the paper (as opposed to the TeX source) are rounded in a way that results in substantial bias in the predictions of the scaling law. In contrast, our re-derivation of the scaling law using the third approach yields results that are compatible with the findings from the first two estimation procedures described by Hoffmann et al.\let\thefootnote\relax\footnotetext{We thank Nuño Sempere, Tom Adamczewski, Eric Michaud, Tony Wang, Nathan Labenz and Jaime Sevilla for their suggestions.}
\let\thefootnote\relax\footnotetext{All our analysis can be replicated using the following address: \url{https://epochai.org/code/analyzing-chinchilla-repo}
}
\end{abstract}

\section{Introduction}

\cite{hoffmann2022training} investigate the optimal model size and number of training tokens for training a transformer language model under a given compute budget. They train over 400 models ranging from 70M to over 16B parameters on 5B to 500B tokens and find that for compute-optimal training, the model size and number of training tokens should scale at equal rates: for every doubling of model size, the number of training tokens should also be doubled. They then train a model called Chinchilla that is compute-optimal according to their results. For this reason, the scaling laws they propose are often called ``Chinchilla scaling laws''.

The authors use three different approaches to estimate the compute-optimal frontier (A1-A3):
\begin{enumerate}[label=A\arabic*]
    \item Training models of fixed sizes on varying numbers of tokens.
    \item Training models of varying sizes targeting fixed compute budgets (IsoFLOP profiles).
    \item Fitting a parametric model of the loss as a function of model size and training tokens.
\end{enumerate}
The key result is that for compute-optimal training, the model size and the number of training tokens should be scaled equally: for every doubling of model size the number of training tokens should also be doubled.

The result that scaling training tokens should grow at roughly the same rate as the number of model parameters has been replicated by others, such as \cite{anil2023palm}. Similarly, \cite{bi2024deepseek} find that training tokens and model parameters should be scaled roughly proportionally, but finds that this is sensitive to the quality of the data, and that a lower token-per-parameter ratio is optimal when training on a higher-quality data.

In addition to informing us about optimal scaling, Approach 3 is of particular interest because it sheds light on the parametric form of the scaling laws for dense transformers. The specific parametric estimates from Hoffmann et al. have been of independent scientific interest, such as in the theoretical explanations of neural scaling laws (e.g. \citealt{michaud2024quantization, bordelon2024dynamical}), and questions about optimizing compute expenditure across both training and inference (e.g. \citealt{sardana2023beyond}). 

We partially reconstruct the dataset from Hoffmann et al. and attempt to replicate Approach 3. This involves fitting a parametric function to model the final pre-training loss as $L(N, D) = E + \frac{A}{N^\alpha} + \frac{B}{D^\beta}$, where $N$ represents the number of model parameters and $D$ represents the number of training tokens. Our analysis reveals that our estimated model differs substantially from the fit reported in Hoffmann et al., and that their fit fails to adequately describe the reconstructed data. We demonstrate that the confidence intervals reported by Hoffmann et al. are implausibly tight and unlikely to be obtained from proper statistical procedures given the size of their dataset. Finally, we finally show that their fit is inconsistent with the scaling policies derived through other approaches, and with the scaling policy suggested by our fit. 

\section{Extracting data from Hoffmann et al.'s Figure 4}

We reconstruct the data from Figure 4 in Hoffmann et al.
\begin{figure}[h!]
    \centering
    \includegraphics[width=0.4\textwidth]{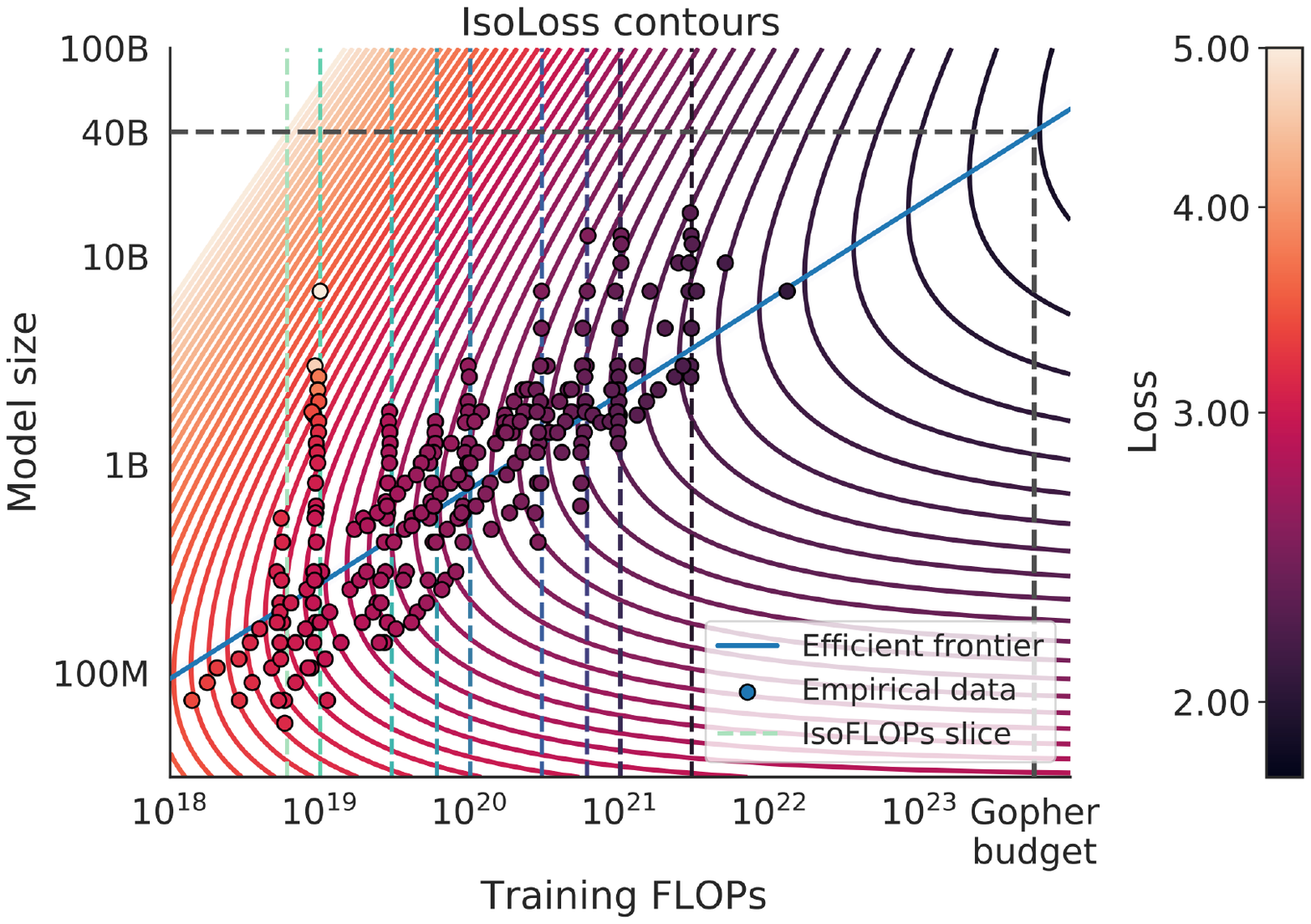}
    \vspace{-0.45cm}
    \caption{\small \centering Display contour from the left sub-figure in Figure 4 taken from Hoffmann et al.}
    \label{fig:fig4_chinchilla}
\end{figure}
Figure 4, which is reproduced in this paper, displays a scatter plot of language models (see Figure \ref{fig:fig4_chinchilla}). The $x$-axis represents model size, the $y$-axis represents training FLOP, and the color of each point encodes the loss value. 
\begin{figure}[h!]
    \centering
    \includegraphics[width=0.5\textwidth]{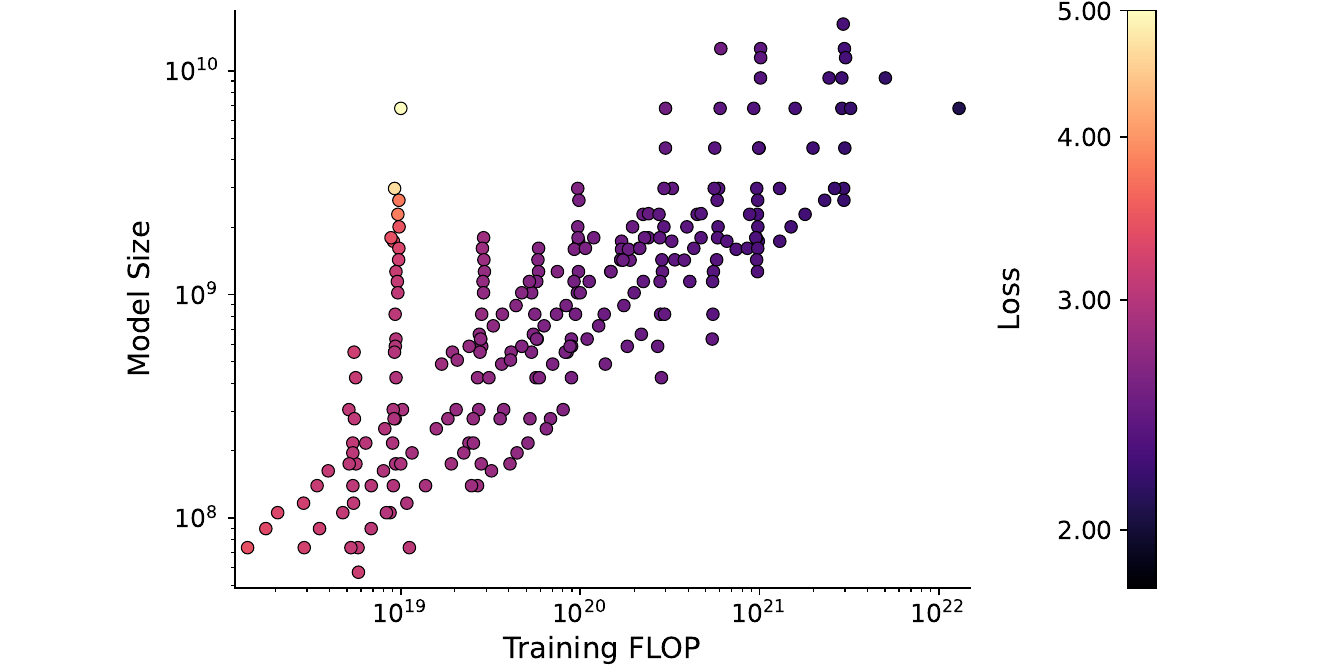}
    \vspace{-0.55cm}
    \caption{\small \centering Our reconstruction of the data from figure 4 in Hoffmann et al.}
    \label{fig:enter-label}
\end{figure}
To extract the data from the figure, we first downloaded the PDF from Hoffmann et al.'s arXiv submission and saved it in SVG format. We then parsed the SVG content to navigate and search the SVG structure. Within the SVG, we identified the group of points representing the scatter plot data and iterated over each point to extract its fill color and position ($x$ and $y$ coordinates) using the attributes of the corresponding SVG elements.

To map the SVG coordinates to the model size and training FLOP values, we used the location of the labels or ticks on the respective axes. This allowed us to establish a correspondence between the SVG coordinates and the actual data values represented in the plot.

To derive the loss values associated with each scatter point, we further extracted the color values (in hex format) from the color scale provided in the graph. The graph's color scale maps colors to loss values using a logarithmic scale ranging from 2.00 to 5.00. We processed the color scale by iterating through the pixels from top to bottom, reading the hex color values from each pixel, and calculating the corresponding loss value based on the pixel's vertical position within the scale. This process resulted in a mapping between hex color values and their corresponding loss values. We then used this mapping to determine the loss value for each scatter point based on its fill color.

The digitization process may introduce some noise or error due to the following factors:

\begin{enumerate}
\item \textit{Imprecise $y$-coordinate extraction}. The $y$-axis of the scatter plot lacks tick marks, making it challenging to determine the exact $y$-coordinates corresponding to specific model sizes. We assumed that the labels indicating the model sizes were positioned such that the vertical center of each text label aligns with its respective value on the $y$-axis.
\item \textit{Precision loss due to loss-to-color mapping}. The color scale is limited to 256 hex values, which restricts our ability to precisely estimate the loss value to an accuracy of approximately 0.01.
\end{enumerate}

A few datapoints stand out. There is a column of points with training FLOP around $10^{19}$ where some of the larger models obtain high loss values (up to 70\% higher compared to models with similar compute budgets). These correspond to the lowest ratio of training tokens to parameters (the top five have a ratio $<0.4$). It is unclear why these training runs yielded such high losses. In our analysis, we will sometimes drop the five points in this column (which we refer to as ``the five outliers" hereafter). In Appendix \ref{sec:what-if}, we show that including these would not materially change our conclusions about the poor fit of Hoffmann et al.'s estimated scaling law.

\section{Hoffmann et al.'s Approach 3 Replication attempt}

We fit the Hoffmann et al. parametric scaling law to our reconstructed subset of the data:
\begin{equation}
    L(N, D) = E + \frac{A}{N^{\alpha}} + \frac{B}{D^{\beta}}
\end{equation}
with $N$ being the number of parameters and $D$ being the number of training tokens. To fit this function used by Hoffmann et al, we minimize the Huber loss:
\begin{equation}
\label{eq:huber-loss}
\small
\min_{a,b,e,\alpha,\beta} \sum_{\text{Run } i} \text{Huber}_\delta \bigg(
\text{LSE}\left( a - \alpha \log N_i, b - \beta \log D_i, e \right) - \log L_i
\bigg),
\end{equation}
where $LSE$ is the log-sum-exp operator.
We follow the methodology in Appendix D.2. in Hoffmann et al. and set $\delta$  to $10^{-3}$, and use a grid of initialisations given by: \(\alpha \in \{0, 0.5, \ldots, 2\}\), \(\beta \in \{0, 0.5, \ldots, 2\}\), \(e \in \{-1, -0.5, \ldots, 1\}\), \(a \in \{0, 5, \ldots, 25\}\), and \(b \in \{0, 5, \ldots, 25\}\). Doing so yields the following estimated model:

\begin{tcolorbox}[colback=green!5!white,colframe=green!60!blue,title=Our estimated model]
\begin{equation}
L(N, D) = 1.8172 + \frac{482.01}{N^{0.3478}} + \frac{2085.43}{D^{0.3658}}
\end{equation}
\end{tcolorbox}

\begin{tcolorbox}[colback=green!5!white,colframe=green!60!blue,title=Hoffmann et al.'s estimated model]
\begin{equation}
\label{eq:hoffmann-law}
L(N, D) = 1.6934 + \frac{406.4}{N^{0.3392}} + \frac{410.7}{D^{0.2849}}
\end{equation}
\end{tcolorbox}

The values we report in Equation \ref{eq:hoffmann-law} for the scaling law parameters in \cite{hoffmann2022training} are more precise than those that may be found in the body of the paper. These more precise estimates are available in the form of comments in the TeX source documents for the paper (which may be found on arXiv). As we will see later, especially the increased precision of the data exponent that we're able to obtain this way makes a big difference to the model's goodness of fit to the data.

We test how significant the differences between our estimates and Hoffmann et al.'s estimates are. This test is based on the following observation: if Hoffmann et al.'s estimates were optimal on the full dataset and our dataset was a generic or random subset of the original dataset they used, then our estimates should approximately follow a normal distribution centered around their estimates with some covariance matrix \( \Sigma \).

\begin{table}[ht]
\small
\centering
\begin{tabular}{@{}lcc@{}}
\toprule
Parameter & Our estimate   & Hoffmann et al's estimate \\ \midrule
$A$ & $\underset{(124.58)}{482.01}$ & 406.4 \\
$B$ & $\underset{(1293.23)}{2085.43}$ & 410.7 \\
$E$ & $\underset{(0.03)}{1.8172}$ & 1.6934 \\
$\alpha$ & $\underset{(0.02)}{0.3478}$ & 0.3392 \\
$\beta$ & $\underset{(0.02)}{0.3658}$ & 0.2849 \\
$a = \beta/(\alpha+\beta) $ & $\underset{(0.02)}{0.5126}$ & 0.454 \\
\midrule
Data points & 240 & $ > 400 $ \\ \bottomrule
\end{tabular}
   \caption[Caption]{\small Our parameter estimates and their standard errors. The standard errors are shown in parentheses and are obtained by bootstrapping. We show the estimate from Hoffmann et al. along with our estimates for comparison.\protect\footnotemark}
\label{tab:param-estimates} 
\end{table}
\footnotetext{We do not use the grid of initializations for our bootstraps, and use the BFGS optimizer rather than Hoffmann et al.'s L-BFGS. We find that using this optimizer enables us to forego having the use a grid of initializations to get very close to the local minimum.}

In other words, if \( \mu \) denotes the Hoffmann et al. estimates and \( \nu \) denotes our best fit, the difference \( \mu - \nu \) should follow \( \mathcal N(0, \Sigma) \) for some covariance matrix \( \Sigma \).  Given that this is the case, we expect \( (\mu - \nu)^T \Sigma^{-1} (\mu - \nu) \) to follow a \( \chi^2 \) distribution with the number of degrees of freedom equal to the number of dimensions of the vectors \( \mu, \nu \). 

The covariance matrix \( \Sigma \) is unknown, but we can estimate it by following the bootstrapping philosophy: ``the bootstrap is to the sample as the sample is to the population". We construct each bootstrap by sampling \( n = 240 \) times with replacement from our full dataset of \( 240 \) data points and then fitting the scaling law to this bootstrapped dataset. Repeating this for \( 4000 \) different bootstraps allows us to obtain a sampling distribution for the vector of parameters \( (\log A, \log B, \log E, \alpha, \beta) \), from which we can compute a sample covariance matrix. Doing this gives us the \( \Sigma \) we need for the \( \chi^2 \) test above, and the $p$-value of the resulting \( \chi^2 \) test ends up being $<10^{-51}$ if we exclude the five outliers from our dataset and $ < 10^{-35} $ if we include them in the dataset when computing \( \Sigma \) using bootstrapping (we always exclude them when computing our parameter estimates). This means that the difference between our parameters and Hoffmann et al.'s parameters is extremely statistically significant.

Examining the parameters individually, $E$ and $\beta$ show highly significant differences, with $p$-values of $ 2.6 \times 10^{-6}$ and $1.1 \times 10^{-4}$, respectively. These small $p$-values indicate that the estimated values for $E$ and $\beta$ are significantly different from their respective true values, suggesting substantial deviations in these parameters' estimates from expected outcomes based on the model. On the other hand, the parameters \textit{A}, \textit{B}, and $\alpha$ have estimated values that are not statistically significantly different from the values reported in Hoffmann et al.

\subsection{Our estimates fit the data better compared to the reported and non-rounded values}
\label{sec:poor_fit}
When plotting the residuals of this fitted scaling law and those of our estimate of the same scaling law, it becomes clear that the estimated scaling law from \cite{hoffmann2022training} with the rounded parameter values reported in the paper ($E = 1.69$, $A = 406.4$, $B = 410.7$, $\alpha = 0.34$, $\beta = 0.28$) fails to fit the data well. 
\begin{figure}[h!]
    \centering
    \includegraphics[width=\linewidth]{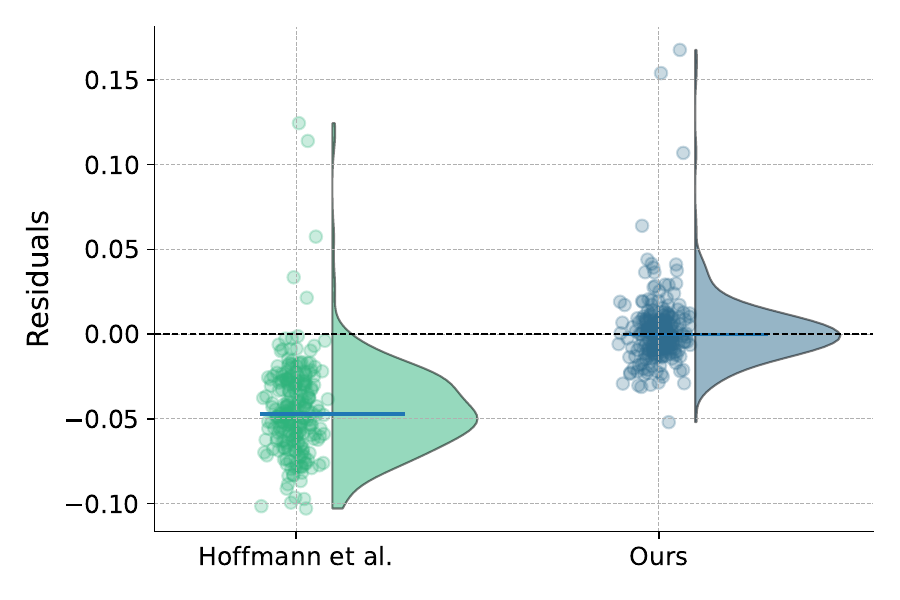}
    \caption{\small Plot of residuals of Hoffmann et al.'s estimated model with rounded parameter values and our estimated model.)}
    \label{fig:rounded-residuals}
\end{figure}
\vspace{-0.3cm}
The residuals for our estimated scaling law are tightly concentrated around 0, indicating a substantially better fit. We find that 98\% of our Huber loss values are smaller than the median loss value of Hoffmann et al, and that our model obtains lower losses for 90\% of all observations. A likelihood ratio test enables us to confidently reject the hypothesis that Hoffman et al.'s estimate  performs as well as our fit ($p < 10^{-235}$, see Appendix \ref{sec:likelihood_ratio}). In other words, if our data reconstruction is correct, the Hoffmann et al. reported scaling law fails to fit the data.

After publishing the first version of this paper, we've been able to ascertain that there are two major reasons behind the worse fit of the Hoffmann et al. parameters to the data:

\begin{enumerate}
    \item The parameter values reported in the paper are rounded, and especially for the data exponent this rounding is significant enough to make the fit much worse than it should be. The true value of the data exponent \( \beta \) should be around \( 0.2849 \) according to the comments in the TeX source code of \cite{hoffmann2022training}; using the value \( 0.28 \) instead introduces a positive bias on the order of \( \approx (10^{11})^{0.0049} - 1 \approx 13 \% \) for a typical dataset size of \( D = 10^{11} \text{ tokens} \) in the finite data component of the reducible loss. This effect is large enough to explain the Hoffmann et al. residuals not ending up centered around zero.

    \item However, even when we use the more precise parameter values from the TeX source, the Hoffmann et al. parameters are still a worse fit to the data (by around 40-50 nats in log likelihood, more on how we compute this later) than our parameters, and this remains true whether or not we exclude the outliers with five highest loss values in the dataset. The authors of \cite{hoffmann2022training} have confirmed this is because they averaged the Huber loss values over different data points instead of summing them during loss minimization and the high loss scale of their optimizer caused their optimization to terminate early, resulting in sub-optimal parameter values.
\end{enumerate}

\begin{figure}[ht]
    \centering
    \includegraphics[width=\linewidth]{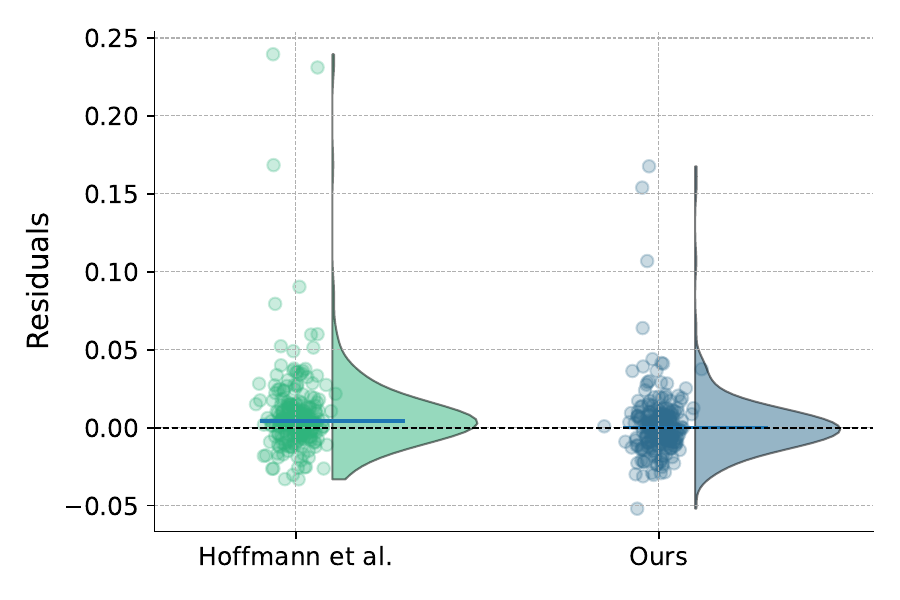}
    \caption{\small Plot of residuals of Hoffmann et al.'s estimated model with unrounded parameter values and our estimated model.)}
    \label{fig:unrounded-residuals}
\end{figure}

We first show the effects of correcting for (1) by using the precise parameter values from the TeX source of \cite{hoffmann2022training} instead of the rounded values from the published paper. The improvement in goodness of fit is substantial, closing most (though not all) of the gap between the two fits in figure \ref{fig:unrounded-residuals}.

The differences no longer look visually striking, so it's worth wondering whether this was the entire problem with the parameter values from \cite{hoffmann2022training}. As we've stated in (2), this turns out to not be the case, and we can precisely quantify the difference in goodness of fit by defining a probability distribution whose negative log likelihood is equal to Huber loss and performing a likelihood ratio test. The results of this are in Table \ref{tab:log-likelihoods}, and the details of how we define the associated probability distributions are available in Appendix \ref{sec:likelihood_ratio} for the interested reader.

\vspace{-0.3cm}

\begin{table}[ht]
\small
\centering
\begin{tabularx}{0.5\textwidth}{@{}Xcc@{}}
\toprule
Parameters & \vtop{\hbox{\strut Log likelihood}\hbox{\strut without outliers}}  & \vtop{\hbox{\strut Log likelihood}\hbox{\strut with outliers}} \\ \midrule
Hoffmann et al.'s fit (with rounding) & $562.25$ & $531.89$ \\
Hoffmann et al.'s fit (without rounding) & $837.78$ & $714.43$ \\
Our best fit & $879.77$ & $757.80$ \\
\midrule
Likelihood ratio test p-value & $1.22\cdot 10^{-16}$ & $ 3.23 \cdot 10^{-17} $\\
Data points & 240 & 245 \\ \bottomrule
\end{tabularx}
\caption{\small The log-likelihood values for the Hoffmann et al. parameters and our best-fit parameters on our dataset, excluding the five outliers. Assuming the unrounded Hoffmann et al. parameters as the null hypothesis, we perform a likelihood ratio test to show that the difference between the log-likelihoods remains highly significant.}
\label{tab:log-likelihoods}
\end{table}

\newpage 

In addition to this, as mentioned earlier, the \( \chi^2 \) test for parameter inequality also remains highly significant (\( p < 10^{-48} \) without the five outliers, \( p < 10^{-35} \) with the five outliers included when computing the covariance matrix \( \Sigma \)): if the true optimal parameters were in fact the unrounded parameters from \cite{hoffmann2022training}, it's unlikely that we would have found parameters as different from them as we have, even taking into account that our \( 240 \) data points might be a random subset of the full dataset of \( > 400 \) data points that \cite{hoffmann2022training} may have used for their fit. All of this is enough to make the case that (1) is not the only reason behind the bad fit of the \cite{hoffmann2022training} parameters to the data.

We make some effort to clarify this point because it might be tempting to assume that since most of the visually apparent evidence of bad fit in Figure \ref{fig:unrounded-residuals} has disappeared upon correcting (1), the early stopping of the optimizer that produced worse fits in \cite{hoffmann2022training} was only a minor problem, at least when it comes to the point estimates of the parameters. This is not true: the difference in goodness of fit remains significant even after correcting for (1) and implies substantial differences in optimal scaling policy when it's extrapolated out to e.g. the compute scale of the Chinchilla 70B model.

\subsection{Hoffmann et al. report implausibly narrow confidence intervals}
\label{sec:confidence_intervals}

A further perplexing feature of Hoffmann et al.'s estimates, once again attributable to the early stopping problem they had with their optimizer, is their extremely narrow confidence intervals for parameters $a$ and $b$, which are defined as:
\begin{equation}
    a \equiv \frac{\beta}{\alpha+ \beta}, \hspace{0.15cm} b \equiv 1-a.
\end{equation}
The significance of these coefficients lies in their role in understanding optimal scaling relationships, specifically how the optimal allocation of a computational budget $C$ can be modeled. The relationships $N_{opt} \propto C^{a}$ and $D_{opt} \propto C^{a}$ use these coefficients to describe how best to allocate training compute.

In particular, they report a confidence intervals of $0.454$ to $0.455$ and $0.542$ to $0.543$ for $a$ and $b$ respectively. These are very tight given that they likely had on the order of $400$ observations with which to estimate $\alpha$ and $\beta$.

By contrast we estimate a standard error of 0.018 for $a$ (See Table \ref{tab:param-estimates}). This would correspond to a width of the 80\% confidence interval of $2 \cdot z_{0.9} \cdot 0.018 \approx 0.05$. Hence our 80\% confidence interval is roughly 50-fold wider than that reported by Hoffmann et al.

How many training runs would we need to observe to get a confidence interval of 0.001? Since the standard errors shrink in $\sqrt{N}$, we would need to increase the number of experiments by a factor of $50^2$ or $2500$. That means that we would need to have access to the results from nearly $240\times 2116 = 600,000$ training runs to obtain a confidence interval as tight as that reported by Hoffmann et al. 

Based on Hoffmann et al.'s report of having ``over 400" observations, we interpret this to mean they likely had between 400 and 500 data points. If, as they claimed, they used only the final loss values for each trained model, it seems unlikely that they would have had hundreds of thousands of observations.

After the publication of the first version of this paper, one of the lead authors of \cite{hoffmann2022training} has clarified that the reason behind the low standard errors was a high loss scale in their L-BFGS-B minimizer resulting from them averaging Huber loss values over examples instead of summing them (\cite{borgeaud2024great}). This caused early termination of the optimization process, both during the original model fit and during bootstrapping. The early stopping of loss minimization during bootstrapping resulted in little movement of the parameter values from their initialization and gave the implausibly narrow confidence intervals that they report for $a$ and $b$.

\subsection{Hoffmann et al.'s Approach 3 scaling policy is inconsistent with Chinchilla and our estimates}

Note that our confidence interval for $a$ is consistent with Approaches 1 and 2 in Hoffmann et al. in the sense that our confidence intervals overlap with theirs. This implies that our optimal scaling policy is consistent with the scaling recommendations from those approaches. 

\begin{figure}[h!]
    \centering
    \includegraphics[width=0.5\textwidth]{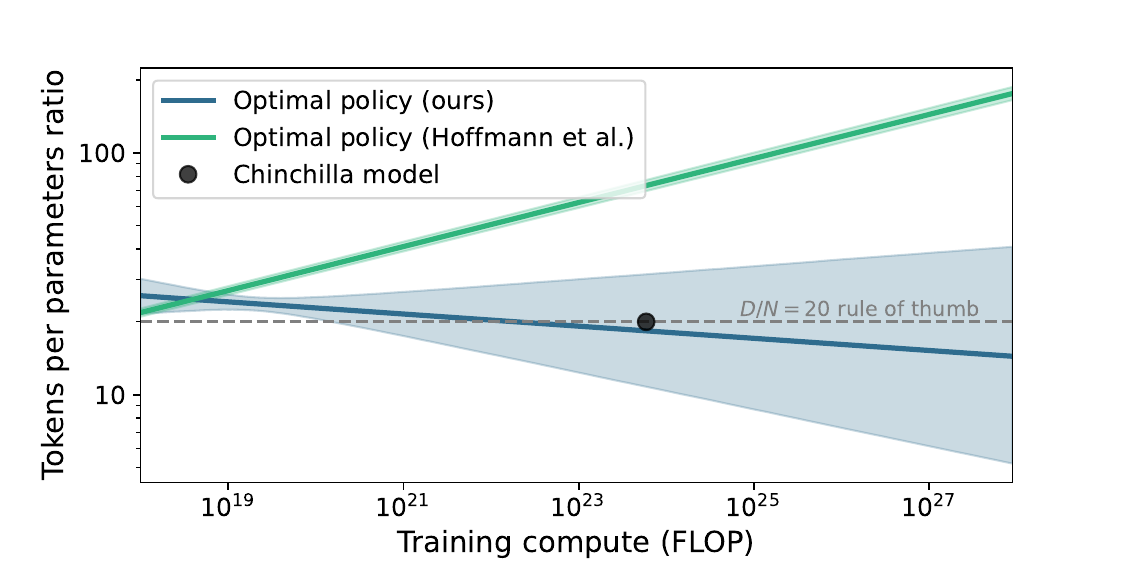}
    \caption{\small Optimal ratio of training tokens to model parameters using our estimates. Shaded regions represent 80\% confidence intervals. While our estimates are consistent with the scaling policy used for Chinchilla, their estimates of their parametric model are not.}
    \label{fig:policies}
\end{figure}

We solve for what the compute-optimal scaling policies are. By doing so, and accounting for the uncertainty in our estimates, we obtain the range of policies illustrated in  Figure \ref{fig:policies}. 

Two key observations emerge from this analysis. First, the confidence intervals for the optimal scaling policies based on Hoffmann et al.'s estimates are extremely narrow. As discussed in Section \ref{sec:confidence_intervals}, we argue that these intervals are unjustifiably tight given the reported number of data points.

Second, the scaling policy derived from Hoffmann et al.'s estimated parameters suggests using approximately 70 tokens per parameter for optimal performance. This prescription is inconsistent with the 20 tokens-per-parameter ratio actually used by Hoffmann et al. to train their 70B Chinchilla model. Interestingly, this 20 tokens-per-parameter ratio aligns closely with the results from the two other approaches (Approaches 1 and 2) reported in their paper.

By contrast, our fitted model implies an optimal ratio of around 20 tokens per parameter, which is consistent with both how the Chinchilla model was trained and the findings from Approaches 1 and 2 in Hoffmann et al. (see Figure \ref{fig:policies}). So, while the inconsistency between the prescriptions based on the estimated scaling law (Approach 3) and the results from Approaches 1 and 2 in Hoffmann et al. might initially appear to be a product of the functional form of the scaling law not fitting the data well, it's in fact an artifact of the early optimizer stopping problem that has produced the inaccurate parameter estimates from \cite{hoffmann2022training} in the first place. The only data points for which the functional form seems to perform badly in our fit are those with extremely few data points compared to their number of parameters, which are the five outliers that we've excluded from the fit that produced our main parameter estimates.

We regard the fact that our parameter and standard error estimates can make Approach 3 consistent with Approaches 1 and 2, both in terms of point estimates and in terms of standard errors, as further evidence that our scaling law parameters are indeed a better fit to the data from \cite{hoffmann2022training}.

\section{Discussion}

We have found three potential issues with Hoffmann et al.'s estimates of the Chinchilla scaling law that rely on Approach 3:
\begin{enumerate}
    \item Their estimated model fits the reconstructed data poorly compared to the best possible estimate of the same functional form. Most of this poor fit is due to the rounding of the parameters reported in the paper, but even after correcting for this the Hoffmann et al. parameters remain significantly worse than the best fit parameters we find.
    \item The confidence are implausibly tight given the number of data points. Obtaining confidence intervals that tight would require many hundreds of thousands of observations, while they likely had only $\sim$400.
    \item Their estimated model implies a scaling policy that is inconsistent with their other approaches and their 20-tokens-per-parameter rule-of-thumb.
\end{enumerate}

According to the authors, all of these issues arise because of a problem with their optimizer which led to early stopping: the optimizer halted before it converged to the true optimal parameter values \cite{borgeaud2024great}. Because we run the optimizer to convergence both during initial estimation and during bootstrapping, both our point estimates and our confidence intervals are more accurate given the data.

Hoffmann et al.'s paper has been highly influential in the language modeling community. Its findings have informed the scaling policies of notable models, such as Google's Gemini suite \citep{team2023gemini}, and have likely guided the development of many other models. Moreover, the specific parameter estimates of the scaling law presented in the paper are of significant scientific interest, such as for machine learning theory. Given the wide-reaching impact of this research, it is crucial for the parameter estimates to be robust and reproducible.

Our work further makes clear that, given the size of the numbers of the number of parameters and tokens, simply reporting scaling law exponents to several significant figures does not provide enough precision to accurately establish the relevant relationship.

Finally, our work further highlights the uncertainty about compute-optimal scaling. In our plotted range of compute-optimal scaling policies (see Figure \ref{fig:policies}), we find that our fitted model is consistent with a ratio of tokens to parameters between 4 and 40 at models train on $1\mathrm{e}26$ FLOP or more. Getting tighter estimates before running experiments of this magnitude would, in expectation, save a meaningful fraction of that amount of compute. 

\appendix
\section{Appendix}

\subsection{What if we don't drop the five outliers?}

\label{sec:what-if}

In the previous analysis, we excluded the five experiments with the lowest ratio of training tokens to parameters, as these outliers were challenging to fit using the Chinchilla scaling law. To ensure the robustness of our findings, we now repeat key parts of the analysis without removing these outliers. This additional analysis reveals that our main conclusions remain unchanged, although the parameter estimates are more uncertain when these outliers are included.

\begin{table}[ht]
\small
\centering
\begin{tabular}{@{}lcc@{}}
\toprule
Parameter & Our estimate & Chinchilla's estimate\\ \midrule
$A$ & $\underset{(145.0)}{463.3}$ & 406.4 \\
$B$ & $\underset{(61650)}{12530}$ & 410.7 \\
$E$ & $\underset{(0.0441)}{1.89}$ & 1.693 \\
$\alpha$ & $\underset{(0.0178)}{0.345}$ & 0.339 \\
$\beta$ & $\underset{(0.0543)}{0.452}$ & 0.285 \\
$a = \beta/(\alpha+\beta)$ & $\underset{(0.032)}{0.512}$ & 0.454 \\
\midrule
Data points & 245 & $ > 400 $ \\ \bottomrule
\end{tabular}
   \caption[Caption]{Our parameter estimates and their standard errors. The standard errors are shown in parentheses and are obtained by bootstrapping. We show the estimate from Hoffmann et al. along with our estimates for comparison.\protect\footnotemark}
\label{tab:param-estimates} 
\end{table}

\paragraph{Equality of models} The overall significance of the model is underscored by an implied $\chi^2$ $p$-value of $6 \times 10^{-50}$, indicating substantial deviations in certain parameter estimates. In detail, $E$ and $\beta$ show highly significant differences from their expected true values, with $p$-values of $2.0 \times 10^{-5}$ and $2.4 \times 10^{-3}$, respectively, highlighting considerable deviations in these parameters' estimates.

\paragraph{Optimal scaling} We find a range consistent with the 20 tokens per parameter rule of thumb. Indeed, our point estimates imply that 25.6 tokens per parameters is optimal.

\subsection{Likelihood ratio test}
\label{sec:likelihood_ratio}

We can further quantify this lack of good fit by recasting the Huber loss minimization problem as a maximum likelihood problem and perform a likelihood ratio test. For any value of \( \delta > 0 \), the function \( p: \mathbb R \to \mathbb R^{\geq 0} \)
\begin{equation}
    p_{\delta}(x) = \frac{\exp(-\text{Huber}_\delta(x))}{\int_{-\infty}^{\infty} \exp(-\text{Huber}_\delta(x)) \, dx}
\end{equation}
is a legitimate probability density function on the real numbers whose negative log-likelihood equals the Huber loss function up to an additive constant.\footnote{It is possible to explicitly compute the integral in the denominator as \( \sqrt{2 \pi} (2 \Phi(\delta) - 1) + 2 e^{-\delta^2/2}/\delta \) where $ \Phi $ is the standard normal distribution's cumulative distribution function.} Consequently, we can convert the loss minimization problem into a likelihood maximization problem for the distribution defined by \( p \). Introducing location and scale parameters \( \mu \) and \( \sigma \), we augment this distribution to
\begin{equation}
    p_{\mu, \sigma, \delta}(x) = \frac{1}{\sigma} \cdot p_{\delta} \left( \frac{x - \mu}{\sigma} \right)
\end{equation}
and then convert the loss minimization problem from Equation \ref{eq:huber-loss} into the following negative log-likelihood minimization problem:
\begin{equation}
\label{eq:huber-likelihood-problem}
\begin{split}
\min_{a,b,e,\alpha,\beta,\sigma} & \sum_{\text{Run } i} -\log p_{\mu=0, \sigma=\sigma, \delta=\delta} \bigg( \\
& \text{LSE}\left( a - \alpha \log N_i, b - \beta \log D_i, e \right) - \log L_i
\bigg). \\
\end{split}
\end{equation}

We perform three fits: one unconstrained fit exactly as described in Equation \ref{eq:huber-likelihood-problem} and two where \( \sigma \) is allowed to vary freely but the other parameters are fixed at the values reported for them in Hoffmann et al. We consider both the rounded parameter values reported in the body of \cite{hoffmann2022training} and the more precise values that may be found in the TeX source.

Then, we compare the two log-likelihoods we obtain using a likelihood ratio test: we assume the Hoffmann et al. parameters as the null hypothesis and report a p-value for the log-likelihood difference under the \( \chi^2 \) distribution with \( 6-1 = 5 \) degrees of freedom, as this is the asymptotic distribution of the test statistic per \cite{wilks1938large}.

\bibliographystyle{apalike} 
\bibliography{bib} 

\end{document}